\documentclass{article}

    \PassOptionsToPackage{numbers, compress}{natbib}
    \usepackage[preprint]{neurips_2023}

\usepackage[utf8]{inputenc} % allow utf-8 input
\usepackage[T1]{fontenc}    % use 8-bit T1 fonts
\usepackage{hyperref}       % hyperlinks
\usepackage{url}            % simple URL typesetting
\usepackage{booktabs}       % professional-quality tables
\usepackage{amsfonts}       % blackboard math symbols
\usepackage{nicefrac}       % compact symbols for 1/2, etc.
\usepackage{microtype}      % microtypography
\usepackage{xcolor}         % colors

\usepackage{graphicx}       % for table control
\usepackage{caption}
\usepackage{algorithm}
\usepackage{algorithmicx}
\usepackage{algpseudocode}
\usepackage{amsmath}
\usepackage{amsthm}
\usepackage{wrapfig}
\usepackage{amssymb}
\usepackage{multirow}
\usepackage{multicol}
\usepackage{enumitem}
\usepackage{caption}
\usepackage{subfigure}
\usepackage{makecell}
\usepackage{pifont}
\usepackage{ragged2e}

\newcommand{\ieno}{\textit{i}.\textit{e}.}

\newcommand{\egno}{\textit{e}.\textit{g}.} %there is no space
\newcommand{\etc}{\textit{etc}.}
\newcommand{\etcno}{\textit{etc}} %there is no "."

\newcommand{\ours}{ResponsibleTA}
\newcommand{\ourfesibility}{DSFP}

\title{Responsible Task Automation: Empowering Large Language Models as Responsible Task Automators}

\author{%
  Zhizheng Zhang\thanks{Equal contribution.}\quad\;
  Xiaoyi Zhang$^{*}$\quad\;
  Wenxuan Xie\quad\;
  Yan Lu\quad\; \\
  Microsoft Research Asia \\ 
  \texttt{\{zhizzhang,\;xiaoyizhang,\;wenxie,\;yanlu\}@microsoft.com}
  }

\begin{document}

\maketitle

\begin{abstract}
  The recent success of Large Language Models (LLMs) signifies an impressive stride towards artificial general intelligence. They have shown a promising prospect in automatically completing tasks upon user instructions, functioning as brain-like coordinators. The associated risks will be revealed as we delegate an increasing number of tasks to machines for automated completion. A big question emerges: how can we make machines behave responsibly when helping humans automate tasks as personal copilots? In this paper, we explore this question in depth from the perspectives of feasibility, completeness and security. In specific, we present Responsible Task Automation (ResponsibleTA) as a fundamental framework to facilitate responsible collaboration between LLM-based coordinators and executors for task automation with three empowered capabilities: 1) predicting the feasibility of the commands for executors; 2) verifying the completeness of executors; 3) enhancing the security (e.g., the protection of users' privacy). We further propose and compare two paradigms for implementing the first two capabilities. One is to leverage the generic knowledge of LLMs themselves via prompt engineering while the other is to adopt domain-specific learnable models. Moreover, we introduce a local memory mechanism for achieving the third capability. We evaluate our proposed ResponsibleTA on UI task automation and hope it could bring more attentions to ensuring LLMs more responsible in diverse scenarios.
  
%   The recent success of Large Language Models (LLMs) signifies an impressive stride towards artificial general intelligence. They have shown a promising prospect in automatically completing tasks upon user instructions, functioning as brain-like coordinators. The associated risks will be revealed as we delegate an increasing number of tasks to machines for automated completion, which imposes more and more challenges in making LLMs behave responsibly. In this paper, we investigate how to enhance the reliability of the collaboration between LLM-based coordinators and executors from the perspectives of feasibility, completeness and security when automatically completing tasks as personal copilots upon users' instructions. To this end, we present Responsible Task Automation (ResponsibleTA) as a fundamental framework that empowers the LLM-based coordinator with three capabilities: 1) predicting the feasibility of the instructions for executors; 2) verifying the completeness of executors; 3) enhancing the security (e.g., the protection of users' privacy). We propose and compare two schemes for the first two capabilities. One is to leverage the general knowledge learned by LLMs themselves via prompt engineering while the other is to adopt domain-specific learnable models. Moreover, we introduce a local memory mechanism for the third capability. We evaluate the capabilities of ResponsibleTA on UI task automation and hope it can serve as a fundamental framework to bring more attentions in ensuring LLMs more responsible.
\end{abstract}

\section{Introduction}

Recent advanced Large Language Models (LLMs) \cite{brown2020language,ChatGPT,GPT4,chowdhery2022palm,chung2022scaling,driess2023palm,touvron2023llama,zhang2022opt} exhibit powerful language understanding, reasoning, generation, generalization and alignment capabilities in a vast suite of real-world scenarios. LLMs acquire generic knowledge on open-domain tasks by scaling up deep learning, which marks a significant milestone in the advancement towards artificial general intelligence. Beyond language tasks, LLMs are empowered with multi-modal perception and generation capabilities through collaboration with domain-specific models \cite{yang2023mm,shen2023hugginggpt,wu2023visual}. They have been revolutionizing the field of task automation
% Robotic Process Automation (RPA)~\cite{hofmann2020robotic,ivanvcic2019robotic,syed2020robotic,van2018robotic} 
by connecting LLMs to various domain-specific models or APIs, in which LLMs serve as brain-like coordinators while domain-specific models or APIs function as action executors \cite{liang2023taskmatrix,vemprala2023chatgpt,driess2023palm}. 

Adopting LLMs to constructing a general-purpose copilot for automating diverse tasks 
% that humans need to handle in their daily life 
is still in an incipient exploration stage. Taking the interaction with UIs as an example, browsing and interacting with various websites and APPs to achieve their diverse intentions (\egno, searching in websites,  online shopping, setting changing, \etc) are almost indispensable for many people's daily life. Some of them require a hierarchical action architecture \cite{bacon2017option,ahn2022can} where a high-level instruction from human known as task-level goal needs to be decomposed into a series of low-level step-wise instructions for actual execution. Such complex multi-step tasks requires each step to be reasonably planned and reliably executed in line with human intentions. This actually poses significant challenges in the compatibility between LLM-based coordinators and their executors without intervention in their training. 
% This actually poses significant challenges in terms of reliability. At the root of such reliability is the compatibility between LLM-based coordinators and their executors without intervention in their training.
Their reliable collaboration requires LLM-based coordinators to be very familiar with the capabilities of executors, and to replan the task execution when necessary upon the command completeness of executors. Existing works \cite{wu2023visual,yang2023mm,shen2023hugginggpt,liang2023taskmatrix} just tell LLMs the powers of executors with model/API descriptions via heavy prompt engineering. However, these descriptions are commonly manually written and summary-like, which are inadequate to describe the executors' capabilities and cannot reflect the execution completeness for LLMs. In addition, LLMs are also possible to have insufficient awareness of the task goals and environments, thus issuing unreasonable instructions to executors.
As an increasing number of tasks are delegated to machines for automated completion, the exposure to risks will grow, which necessitates addressing the reliability issue with a sense of urgency.
% These will expose more and more risks as an increasing number of tasks are delegated to machines for automated completion, imposing urgent demands on addressing the reliability issue. 

In this work, we propose \ours~as a fundamental multi-modal framework to empower LLMs as responsible task automators in three aspects: 1) \textit{\textbf{Feasibility}}: \ours~predicts the feasibility of low-level commands generated by the LLM-based coordinator and return the results to LLMs for replanning before action execution. This capability aims to minimize the risks and time consumption of having the executors perform unachievable instructions, making task automation more controllable and efficient. 2) \textit{\textbf{Completeness}}: \ours~checks the execution results of low-level commands step-by-step and provides feedbacks timely for the LLM-based coordinator to allow it reschedule next steps more reasonably. This capability can improve the success rates of automated task completion. 3) \textit{\textbf{Security}}: \ours~augments LLMs with edge-deployed memories, which allows user-sensitive information are hidden during the interaction with cloud-deployed LLMs and are only stored and used locally on users' devices. 
% in which LLMs' commands for executors only contain the queries corresponding to users' private information while their specific contents are only stored locally on users' devices, \ieno, edge-deployed memories.
This capability reduces or avoids the transmission of users' sensitive information between the cloud-deployed LLMs and edge-deployed executors, thus lowering the leakage risk of users' privacy. Empowering LLMs with these
% three
capabilities, \ours~achieves automatic verification before and after each execution step, which improves not only the success rate but also the completion efficiency with thoughtful security guarantees for users.
% efficiency of automated task completion with more thoughtful security guarantees for users.

In addition to the framework design of \ours, we investigate the implementations of three core functionalities of \ours~as an empirical study in a practical application scenario (\ieno, UI task automation). Its goal is to automatically ground target UI elements and interact with them via automatic clicking and typing operations based on task instructions from users, which is of high demands in both academia and industry. For feasibility and completeness, we propose and compare two technical paradigms for their implementations. One paradigm is to leverage generic knowledge of LLMs themselves via prompt engineering.
% with the help of a domain-specific perception model. We train a screen parsing model to translate screenshots into structured linguistic form as the inputs of LLMs and achieve feasibility prediction and completeness verification by LLMs via prompt engineering. 
The other paradigm is to train domain-specific models specifically responsible for these two functionalities as external assistants of the coordinators and executors. We observe that leveraging LLMs themselves is inferior to adopting domain-specific models, demonstrating that domain-specific knowledge is crucial for enhancing the reliability of LLMs in task automation. For security, we introduce a local memory and propose a mechanism design to use it for endowing \ours~with security guarantees.
% To showcase the effectiveness of our proposed framework and provide convincing empirical results, we choose UI task automation as a domain-specific application scenario for our experiments.

The contributions of this work can be summarized as follows:
\begin{itemize}[leftmargin=*,noitemsep,nolistsep]
    \item We present a fundamental multi-modal framework, dubbed as \ours, which empowers LLMs with the capabilities of feasibility prediction, completeness verification and security protection for automatically completing tasks in a responsible manner.
	\item We propose and compare two technical paradigms for implementing the feasibility prediction and completeness verification functionalities of \ours~in UI task automation scenarios. One is to leverage internal knowledge of LLMs themselves while the other is to train domain-specific models as external assistants.
    \item We introduce a local memory mechanism to endow \ours~with the capability of security protection for user privacy, and showcase its effectiveness in some representative practical cases.
\end{itemize}

% \vspace{-2mm}
\section{Related Works}
% \vspace{-2mm}
\subsection{Development of Large Language Models}

Starting from the language model Bert~\cite{devlin2019bert}, the pretraining-finetuning scheme is a common practice in many natural language processing (NLP) and computer vision (CV) problems. By bringing data and models to a larger scale, GPT-3~\cite{brown2020language} demonstrates that LLMs obtain the ability of in-context learning, where LLMs can quickly adapt to a new task given only a few examples as prompts. Furthermore, InstructGPT~\cite{ouyang2022training} finetunes GPT-3 with a dataset of human
% labeler 
demonstrations to make LLMs follow users' intents. Recently, ChatGPT~\cite{ChatGPT} and GPT-4~\cite{GPT4} demontrate excellent ability in generating complete answers to users' natural language questions. Considering that these LLMs only deal with language tokens, Kosmos-1~\cite{huang2023language} extends its training set to image-text data and shows capability in visual question answering and multimodal dialog. In the meanwhile, there are concurrent LLMs such as PaLM~\cite{chowdhery2022palm,chung2022scaling} and PaLM-E~\cite{driess2023palm}, and open-source efforts such as LLaMA~\cite{touvron2023llama} and OPT~\cite{zhang2022opt}. Despite the strong performance, answers generated by LLMs are not always reliable. Our work aims to empower LLMs to be reliable in the area of task automation.
% , in which it is critical to avoid erroneous executions.

\subsection{Large Language Models for Task Automation}

LLMs are able to serve as actors or coordinators/planners for digital or physical AI task automation. When functioning as actors, the outputs of LLMs are executable actions \cite{micheli2021language,ahn2022can}, which is limited to natural language processing tasks. Towards broader applications in physical \cite{huang2022language,ahn2022can,lin2022grounded,driess2023palm,liang2023taskmatrix}, simulated physical \cite{volum2022craft,wang2023describe} and digital \cite{shen2023hugginggpt,wu2023visual,yang2023mm,liang2023taskmatrix} environments, LLMs usually act as a brain-like coordinator to process human high-level instructions into step-wise executable machine commands and hand them over to domain-specific models/APIs for actual execution. Along this route, LLMs have been opening up infinite possibilities for task automation and putting forward higher requirements for the reliability of automated systems meanwhile. The knowledge defects or biases of LLMs and the gap between LLMs and executors both possibly lead to potential risks in the failure of task completion or even causing harm to users \cite{greshake2023llmsecurity}. Previous work \cite{ahn2022can} attempts to address the executability issue by training an external model to critic the outputs of LLM. It delivered success in simple robotic environments \cite{ahn2022can} but is demonstrated hard to be applied to complex environments with more objects and diverse actions \cite{shridhar2020alfworld,wu2023plan}. Besides, \cite{wang2023describe,wu2023plan} track the execution results for the follow-up planning of LLMs. To the best of our knowledge, we are the first to systematically study
% take into account 
the reliability of LLM-based task automation from feasibility, completeness and security perspectives.

\begin{figure}[t]
	\begin{center}
		\includegraphics[scale=0.4]{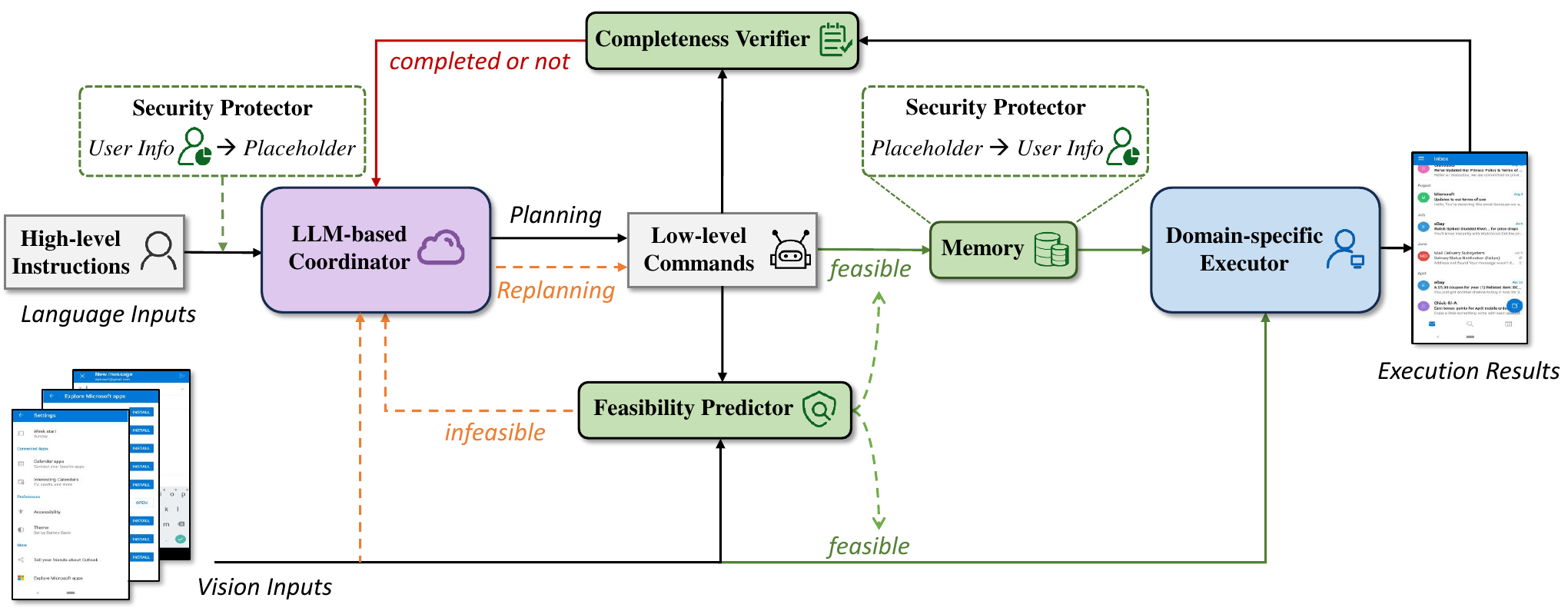} 
	\end{center}
	\caption{The framework of the proposed \ours. It augments the cloud-deployed LLM-based coordinator with feasibility protector, completeness verifier and a local memory, achieving its responsible collaboration with the domain-specific executor. They are all detailed in the main text.}
	\label{fig:framework}
\end{figure}

\section{Method}

% In this section, we first present the framework design of \ours, and then introduce the technical paradigms for implementing the core functional modules in \ours.

\subsection{\ours~Framework}

\ours~is a fundamental multi-modal framework for empowering LLMs as responsible task automators. The goal is to comprehensively enhance the reliability of employing LLMs as coordinators, while adopting domain-specific models/APIs as executors, for automatically completing tasks in according with human instructions. 
% when using LLMs as coordinators while domain-specific models/APIs as executors for automatically completing tasks in according with human instructions.
It achieves reliability enhancement with three empowered capabilities: feasibility prediction, completeness verification and security protection.

% \ours~can be generally applied to a broad range of task automation in physical or digital environments. 
As shown in Figure \ref{fig:framework}, \ours~takes high-level instructions (\ieno, task goals) from human as inputs and parses them into low-level (\ieno, step-wise) commands with a pretrained LLM for action planning.
% task planning.
% by leveraging its generic procedural knowledge about the open-ended world for task planning. 
To ensure the low-level command is reasonably generated and correctly executed at each step, we introduce a feasibility predictor and a completeness verifier in \ours. Before each execution, the feasibility predictor assesses whether the generated low-level command is executable in practice upon the current screenshot and this command itself. 
% The feasibility predictor assesses whether the low-level command generated by the LLM-based coordinator is executable in practice for a given executor.
It will ask the LLM-based coordinator for a replanning via prompt engineering when the command is judged as infeasible, otherwise feed the command to the executor when feasible.
% once feasible. 
When reaching the pre-set maximum number of replanning attempts, the LLM will terminate the current task and offer feedbacks for users. 
% When reaching the pre-set maximum number of replanning attempts, the LLM will provide the user with the results of the feasibility predictor and request more detailed instructions.
After each execution, the completeness verifier checks whether the execution result aligns with the goal of given low-level command. Replanning will be launched by LLM once an invalid or unfulfilled execution is detected. Note that we input the UI element information on the screen to the LLM in linguistic form through a trained screen parsing model only when replanning is needed, so as to protect personal information on users' screens as much as possible. (See supplementary for more details.) Moreover, we integrate a security protector into \ours~to allow user-sensitive information to be stored locally. With this module, users are allowed to replace all sensitive information by their predefined placeholders, and translates them back when sending the commands to edge-deployed executors for local execution. 
% Users are allowed to replace all sensitive information that they do not want to upload to the cloud with their predefined placeholders. The security protector takes these placeholders in low-level commands as queries and translates them back to real information when sending the commands to edge-deployed executors.
As a result, \ours~enhances the internal alignment between the coordinator and executors, and provides security guarantees for users as a responsible task automation framework.

We employ UI task automation as a practical application scenario for in-depth investigation and experiments, so as to 1) showcase the effectiveness of \ours~intuitively and 2) state the implementations of three core modules in \ours~clearly. 
% In \ours, the domain-specific executor determines specific application scenarios. To 1) showcase the effectiveness of \ours~intuitively and 2) state the implementations of three core modules in \ours~clearly, in this work, we employ UI task automation as a practical application scenario for in-depth investigation and experiments. 
Its goal is to automate the interaction tasks between human and UIs (\egno, searching in websites, playing media, online shopping, changing settings, \etc). We train a domain-specific executor to automatically ground the target UI element by predicting
% as
its spatial coordinates for clicking or typing,
% information inputting 
with a given low-level command and the corresponding screenshot as the inputs. Since this is not the emphasis in this work, we place the detailed model design and training of the executor in the experiment section and supplementary.

\subsection{Feasibility Predictor}
\label{sec:feasibility_predictor}

The feasibility predictor takes the low-level command and the current screenshot as inputs to predict the feasibility of this command. It contributes to avoiding the execution of a infeasible or dangerous command by asking for replanning once infeasible. Here, we introduce two technical paradigms for its implementation, and compare their effectiveness in the following experiment section.

\begin{figure}[t]%[h]
	\begin{center}
		\includegraphics[width=0.95\textwidth]{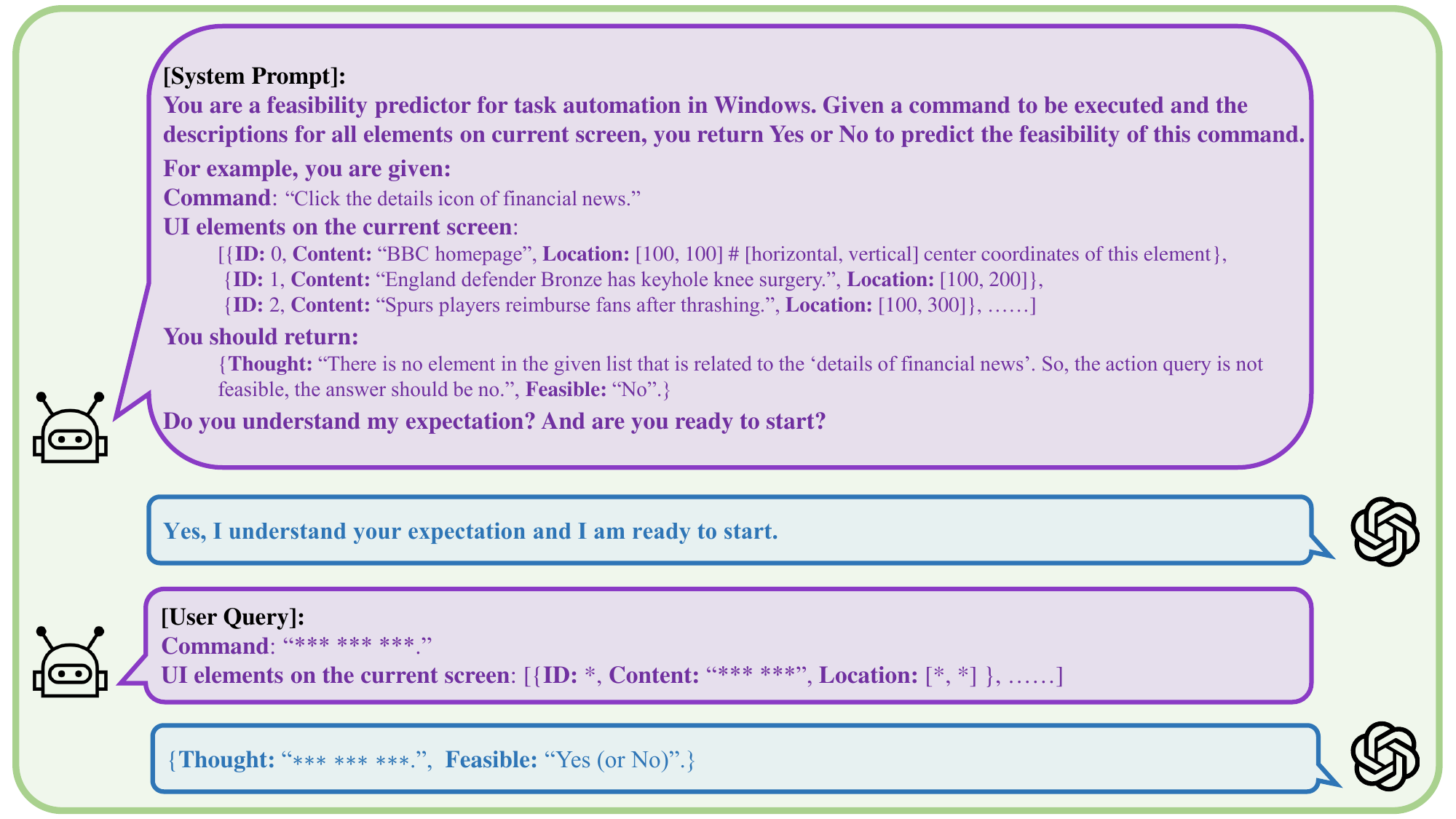} 
	\end{center}
	\caption{Illustration of our prompt engineering based paradigm for implementing the feasibility predictor in our proposed \ours.}
	\label{fig:model_llmfp}
\end{figure}

\paragraph{Prompt engineering based paradigm.} This paradigm is designed to leverage the internal generic knowledge of a LLM for achieving feasibility prediction. We first design a system prompt as the launching prompt to tell LLM our task goal and expected outputs together with some specific demonstrations. Considering that most of advanced LLMs (\egno, ChatGPT, GPT-4) have not integrated or released their input interfaces for other modalities (\egno, vision) except for language, we train a screen parsing model to translate the screenshots into a series of UI elements where each element is represented in linguistic description with its \textit{index}, \textit{text content}, \textit{location} and \textit{type}. More details of this model can be found in Sec. \ref{sec:scree_parser} of the supplementary material. We organize the linguistic descriptions of all UI elements at the current screenshot as a structured form (\ieno, dictionary), and input it together with the low-level command at the current step as the user queries for each task instance. The system prompt and the demonstrations of user queries of our proposed feasibility predictor are illustrated in Figure \ref{fig:model_llmfp} in detail.

\begin{wrapfigure}{r}{0.47\textwidth}
    \vspace{-8mm}
	\begin{center}
		\includegraphics[scale=0.43]{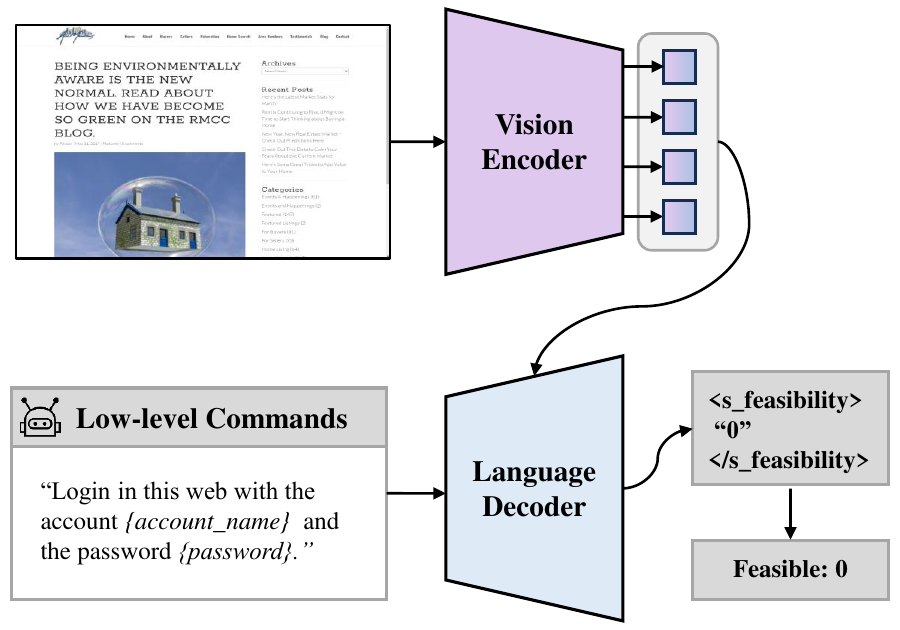} 
	\end{center}
	\vspace{-8pt}
	\caption{Illustration of our domain-specific model based paradigm for implementing the feasibility predictor in \ours.}
	\vspace{-6mm}
	\label{fig:model_dsfp}
\end{wrapfigure}

\paragraph{Domain-specific model based paradigm.} This paradigm aims to address the feasibility prediction by training a domain-specific model, \ieno, an external expert. Here, we devise an end-to-end multi-modal model with its architecture inspired by the Pix2Seq modelling idea \cite{chen2021pix2seq}. As shown in Figure \ref{fig:model_dsfp}, our Domain-Specific Feasibility Predictor, referred to as \ourfesibility, consists of a transformer-based vision encoder and a transformer-based language decoder. Given an image $\mathbf{X}\in \mathbb{R}^{H \times W \times C}$, the vision encoder captures the features of $\mathbf{X}$ and embeds it into $n_v$ $d$-dimensional tokens denoted by $\{\mathbf{v}_i | \mathbf{v}_i \in \mathbb{R}^d, 1 \leq i \leq n_v\}$. The low-level command is tokenized to be $n_t$ $d$-dimensional text tokens denoted by $\{\mathbf{t}_i | \mathbf{t}_i \in \mathbb{R}^d, 1 \leq i \leq n_t\}$. Then the sets $\{ \mathbf{v} \}$ and $\{ \mathbf{t} \}$ are fed into the language decoder to generate the prediction results in linguistic form. 

To make the above architecture design unified to other tasks, such as completeness verification introduced subsequently, we model the outputs of its decoder in a structured linguistic form as ``\textit{<task\_prompt> \{results\} </task\_prompt>}'' following \cite{kim2022ocr,yang2022unitab}. Here, ``\textit{<task\_prompt>}'' and ``\textit{</task\_prompt>}'' denotes the start and end in the linguistic sequence for the results, respectively. And ``\textit{\{results\}}'' denotes the actual contents of the results. For the feasibility predictor of \ours, ``\textit{<task\_prompt>}'' and ``\textit{</task\_prompt>}'' are instantiated to be ``\textit{<s\_feasibility>}'' and ``\textit{</s\_feasibility>}'', respectively. The ``\textit{\{results\}}'' could be 0 or 1 where 0 represents ``\textit{infeasible}'' while 1 denotes ``\textit{feasible}''.

\subsection{Completeness Verifier}
\label{sec:completeness_verifier}

The completeness verifier in \ours~takes the executed command and the screenshot after execution as its inputs to assess the completeness of this command. It in fact plays the role of handling the internal alignment between the LLM-based coordinator and the executor to ensure the reliability of their collaboration. For each step, the verifier provides a feedback for the LLM-based coordinator to tell whether the execution result of the executor aligns with its expectation at this step so that it can remedy erroneous or biased execution timely via a replanning operation. It forms a loop to endow \ours~with the capability of self-correction. We showcase its effectiveness in the case study in the experiment section.

We also propose and compare two technical paradigms for implementing this module, which are similar with those of the feasibility predictor introduced as above. For conciseness, we only state the differences with the feasibility predictor here and place more details in our supplementary. For the prompt engineering based paradigm, we share the screen parsing model with that in the feasibility predictor, and update the system prompt and user queries in line with the goal of completeness verification. Their contents are detailed in the supplementary. For the domain-specific model based paradigm, we adopt the same model architecture and the output format with those of the feasibility predictor, but feed the executed command and the screenshot after execution as its inputs. Here, ``\textit{<task\_prompt>}'' and ``\textit{</task\_prompt>}'' are instantiated to be ``\textit{<s\_completeness>}'' and ``\textit{</s\_completeness>}'', respectively. The ``\textit{\{results\}}'' could be 0 or 1 where 0 represents ``\textit{incomplete}'' while 1 denotes ``\textit{complete}''.

\subsection{Security Protector}
\label{sec:security_protector}

% In task automation scenarios, 
Many tasks that users need to be automated involve their private information. For example, online shopping or computer configuration may require the credit card, account and password information of users. The security protector in \ours~is designed to enable the sensitive information from users to be stored and used locally, obviating the transmission of them to the cloud-deployed LLM and reducing information leaky risks. To this end, the security protector uses a NER model (\egno, Bert-NER \cite{bert_NER}) to automatically detect the sensitive information in users' instructions, and replace them with a information placeholder denoted by ``\textit{\{info\_name\}}''.
% To this end, the users could use information placeholders instead of their specific personal information in their high-level task instructions. As a result, in both high-levels instructions send to LLMs and low-level commands generated by LLMs, the users' private information is allowed to be replaced with a information placeholder denoted by ``\textit{\{info\_name\}}''.
Then, it stores ``\textit{\{info\_name\}}'' and its corresponding real contents using a dictionary structure into an edge-deployed memory. When received a command from LLM, the security protector translates the information placeholder back to its original contents, then sends the translated command to executors for local execution. For example, when users ask \ours~to change the shipping address in a shopping website, their old and new address are represented as ``\textit{\{old\_address\}}'' and ``\textit{\{new\_address\}}'', respectively, when interacting with the cloud-deployed LLMs, and are translated back for local executors.

% in both high-level instructions as well as low-level commands, and are translated into their corresponding contents by the security protector for actual execution in local.

\section{Experiments}

\subsection{Datasets and Implementation Details}

\paragraph{Feasibility prediction dataset.}
We collect data for feasibility prediction from Common Crawl\footnote{\url{https://commoncrawl.org/the-data/}}, an open repository of web crawl data. We sample about 40K web pages from this open repository, yielding 1.1M image-text pairs. Among them, 1M pairs are used for training while the remaining 100K pairs are for testing. For each web page, we extract leaf elements and assign captions to them from HTML entries such as \textit{inner-text}, \textit{value}, \textit{alt}, \textit{aria-label}, \textit{label-for}, and \textit{placeholder}. Subsequently, we employ a randomized algorithm to generate low-level commands for elements by leveraging their captions. Example commands include ``select the \textit{\{element\_caption\}} item'', ``click the item to the right of \textit{\{element\_caption\}}'', ``enter \textit{\{words\}} into \textit{\{element\_caption\}}'', ``scroll until \textit{\{element\_caption\}}'', \etcno. All of these commands are deemed as feasible. Infeasible commands are generated based on fake element captions that do not appear in the current web page.
% The feasibility prediction dataset comes from Common Crawl\footnote{\url{https://commoncrawl.org/the-data/}}, a large-scale open dataset of crawled web pages. We sample a subset from Common Crawl, retain only English pages, and render them into screenshots. As a consequence, a dataset of 1.15 million web pages with screenshots is obtained. For each web page, we extract leaf elements and assign captions to them from HTML entries such as \textit{inner-text}, \textit{value}, \textit{alt}, \textit{aria-label}, \textit{label-for}, and \textit{placeholder}. Subsequently, we employ a randomized algorithm to generate low-level instructions for elements by leveraging their captions. Example instructions include ``select the <element\_caption> item'', ``click the icon at the top left corner'', ``enter <words> into <element\_caption>'', ``scroll until <element\_caption>'', and ``click the item to the right of <element\_caption>''. All of these instructions are deemed as feasible. Infeasible instructions are generated based on fake element captions that do not appear in the current web page.

\paragraph{Completeness verification dataset.}

This dataset is also based on publicly available web pages. This dataset leverages the transitions between two web pages, i.e., it jumps to page B by clicking an element in page A. Based on the transitions we are able to define a positive example of completeness as a 3-tuple [screenshot A, \textit{\{element\_caption\}}, screenshot B]. We generate negative examples by substituting the 3-tuple with fake items. Overall, this dataset includes 113K web pages with 1.2M image-text pairs for training and 6K web pages with 60K image-text pairs for testing.
% The completeness verification dataset is based on publicly available web pages. In addition to the feasibility prediction dataset, the completeness verification dataset also stores the transitions between two web pages, i.e., it jumps to page B by clicking some element in page A. Based on the transitions we are able to define a positive example of completeness as a 3-tuple [screenshot A, <element\_caption>, screenshot B]. We can also generate negative examples by substituting the 3-tuple with fake items.

The data for task automation executor is based on the feasibility prediction dataset. We place more details about the dataset configuration and model implementation in our supplementary.

% \paragraph{Task automation execution.}

\subsection{Quantitative Results}

% Table generated by Excel2LaTeX from sheet 'ALLinOne'
\begin{table}[t]%[htbp]
  \centering
  \caption{The results of quantitative evaluation for the proposed feasibility prediction and completeness verification modules. ``AP'', ``Acc'' and ``F1'' are short for the average precision, accuracy and F1 score metrics, respectively. The superscript ``$+$'' denotes the result evaluated on a larger test dataset.}
    \begin{tabular}{lcccccc}
    \toprule
    \multicolumn{1}{c}{\multirow{2}[0]{*}{\textbf{Models}}} & \multicolumn{3}{c}{\textbf{Feasibility Prediction}} & \multicolumn{3}{c}{\textbf{Completeness Verification}} \\
           & \quad Acc (\%)\quad & \quad AP\quad & \quad F1\quad & \quad Acc (\%)\quad & \quad AP\quad & \quad F1\quad \\
    \hline
    \specialrule{0em}{1.5pt}{1pt}   % control space gap
    LLM-based (ChatGPT)\quad & \quad65.7\quad & \quad0.631\quad & \quad0.546\quad & \quad61.1\quad & \quad0.564\quad & \quad0.704\quad \\
    LLM-based (GPT-4)\quad & \quad68.9\quad & \quad0.669\quad & \quad0.583\quad & \quad62.9\quad & \quad0.575\quad & \quad0.721\quad \\
    DSM-based\quad  & \quad74.8\quad & \quad0.818\quad & \quad0.671\quad & \quad83.8\quad & \quad0.803\quad & \quad0.833\quad \\
    DSM-based$^{+}$\quad & \quad75.3\quad & \quad0.823\quad & \quad0.678\quad & \quad83.5\quad & \quad0.804\quad & \quad0.829\quad \\
    \bottomrule
    \end{tabular}%
  \label{tab:quantitative}%
\end{table}%

We evaluate afore-introduced two technical paradigms for implementing the feasibility predictor and completeness verifier in \ours. 
% They are introduced in Sec. \ref{sec:feasibility_predictor} and Sec.\ref{sec:completeness_verifier} in detail. 
The experiment results are in Table \ref{tab:quantitative}. Due to the heavy cost and efficiency of the testing with LLMs, we randomly sample 5K image-text pairs from the test splits on feasibility prediction and completeness verification for evaluating the prompt engineering based paradigms. As for domain-specific models based paradigms, we evaluate them on the same 5K image-text pairs (abbreviated as DSM in Table \ref{tab:quantitative}) as well as on the entire test splits (abbreviated as DSM$^{+}$ in Table \ref{tab:quantitative}). As shown in Table \ref{tab:quantitative}, we can find that \textit{DSM-based} and \textit{DSM-based$^{+}$} perform very closely under these two test settings. This indicates our sampled 5K image-text pairs are diverse and representative enough for providing convincing evaluation results.

\paragraph{Accuracy of feasibility prediction.} As shown in Table \ref{tab:quantitative}, GPT-4 is more powerful than ChatGPT for feasibility prediction. And the \textit{DSM-based} feasibility predictor outperforms the \textit{LLM-based (GPT-4)} one by 5.9\%, 0.149, 0.088 in accuracy, average precision and F1 score, respectively, on feasibility prediction. These results demonstrate the superior performance of the domain-specific models based paradigm relative to prompt engineering based paradigm on feasibility prediction.

\paragraph{Accuracy of completeness verification.} From the in Table \ref{tab:quantitative}, we observe that \textit{DSM-based} feasibility predictor outperforms the \textit{LLM-based (GPT-4)} one by 20.9\%, 0.228, 0.112 in accuracy, average precision and F1 score, respectively, on completeness verification, showing a similar tendency with that on feasibility prediction. It also indicates the performance superority of adopting a domain-specific model as the completeness verifier in \ours.

As above, we can find the domain-specific model based paradigms perform consistently better than prompt engineering based paradigms on implementing these two functionalities of \ours. Despite this, the advantage of prompt engineering based paradigms is that they do not require the collection of specific data for training, offering better flexibility in practical deployment.

% Table generated by Excel2LaTeX from sheet 'CaseStudy'
\begin{table}[t]%[htbp]
  \centering
  \caption{The case study results of task automation in the real world. The ``\textit{Baseline}'' denotes the model consisting of a LLM-based coordinator and a task automation executor, without our proposed feasibility prediction and completeness verification modules. The ``\textit{+Fea.}'' and ``\textit{+Com.}'' refers to adding the feasibility prediction module and adding the completeness module, respectively. We represent each execution result with its ``\textit{Progress}'' and ``\textit{End Status}''. Here, the ``\textit{Progress}'' is shown in the form of the number of ``\textit{valid steps / total execution steps (human expert steps)}'', in which ``\textit{human expert steps}'' refers to the step number of completing given instruction by a human expert. For the ``\textit{End Status}'', \ding{51} means the task goal has been achieved in the end while \ding{55} indicates it has not. }
  \resizebox{\textwidth}{!}{
    \begin{tabular}{cp{5.5cm}cccccc}
    \toprule
    \multirow{2}[0]{*}{\textbf{No.}} & \multirow{2}[0]{*}{\textbf{High-level Instruction}} & \multicolumn{2}{c}{\textbf{Baseline}} & \multicolumn{2}{c}{\textbf{Baseline+Fea.}} & \multicolumn{2}{c}{\textbf{Baseline+Fea.+Com.}} \\
          &       & Progress & End Status & Progress & End Status & Progress & End Status \\
    \hline\specialrule{0em}{1.5pt}{1pt}
    \multirow{2}[0]{*}{1} & \multirow{2}[0]{*}{Open football news in bbc.com.} & \multirow{2}[0]{*}{4/4 (4)} & \multirow{2}[0]{*}{\ding{51}} & \multirow{2}[0]{*}{4/4 (4)} & \multirow{2}[0]{*}{\ding{51}} & \multirow{2}[0]{*}{4/4 (4)} & \multirow{2}[0]{*}{\ding{51}} \\
          &       &       &       &       &       &       &  \\
    \hline\specialrule{0em}{1.5pt}{1pt}
    \multirow{2}[0]{*}{2} & \multirow{2}[0]{*}{Find the view setting page in Outlook.} & \multirow{2}[0]{*}{4/4 (4)} & \multirow{2}[0]{*}{\ding{51}} & \multirow{2}[0]{*}{4/4 (4)} & \multirow{2}[0]{*}{\ding{51}} & \multirow{2}[0]{*}{4/4 (4)} & \multirow{2}[0]{*}{\ding{51}} \\
          &       &       &       &       &       &       &  \\
    \hline\specialrule{0em}{1.5pt}{1pt}
    \multirow{2}[0]{*}{3} & Navigate to the language setting in my Windows11. & \multirow{2}[0]{*}{3/3 (3)} & \multirow{2}[0]{*}{\ding{51}} & \multirow{2}[0]{*}{3/3 (3)} & \multirow{2}[0]{*}{\ding{51}} & \multirow{2}[0]{*}{3/3 (3)} & \multirow{2}[0]{*}{\ding{51}} \\
    \hline\specialrule{0em}{1.5pt}{1pt}
    \multirow{2}[0]{*}{4} & Find the system setting for text size in my Windows11. & \multirow{2}[0]{*}{1/3 (3)} & \multirow{2}[0]{*}{\ding{55}} & \multirow{2}[0]{*}{3/3 (3)} & \multirow{2}[0]{*}{\ding{51}} & \multirow{2}[0]{*}{3/3 (3)} & \multirow{2}[0]{*}{\ding{51}} \\
    \hline\specialrule{0em}{1.5pt}{1pt}
    \multirow{2}[0]{*}{5} & Help me open the latest received e-mail in my Outlook. & \multirow{2}[0]{*}{2/3 (3)} & \multirow{2}[0]{*}{\ding{55}} & \multirow{2}[0]{*}{3/3 (3)} & \multirow{2}[0]{*}{\ding{51}} & \multirow{2}[0]{*}{3/3 (3)} & \multirow{2}[0]{*}{\ding{51}} \\
    \hline\specialrule{0em}{1.5pt}{1pt}
    \multirow{2}[0]{*}{6} & Go to github.com and check issues that mentioned me, already logged in. & \multirow{2}[0]{*}{2/5 (4)} & \multirow{2}[0]{*}{\ding{55}} & \multirow{2}[0]{*}{4/4 (4)} & \multirow{2}[0]{*}{\ding{51}} & \multirow{2}[0]{*}{4/5 (4)} & \multirow{2}[0]{*}{\ding{51}} \\
    \hline\specialrule{0em}{1.5pt}{1pt}
    \multirow{2}[0]{*}{7} & Log in Instacart with username \{username\} and password \{password\}. & \multirow{2}[0]{*}{5/6 (6)} & \multirow{2}[0]{*}{\ding{55}} & \multirow{2}[0]{*}{6/6 (6)} & \multirow{2}[0]{*}{\ding{51}} & \multirow{2}[0]{*}{6/6 (6)} & \multirow{2}[0]{*}{\ding{51}} \\
    \hline\specialrule{0em}{1.5pt}{1pt}
    \multirow{2}[0]{*}{8} & Go to Amazon and add a pair of gloves into the shopping cart. & \multirow{2}[0]{*}{4/6 (6)} & \multirow{2}[0]{*}{\ding{55}} & \multirow{2}[0]{*}{6/6 (6)} & \multirow{2}[0]{*}{\ding{51}} & \multirow{2}[0]{*}{6/6 (6)} & \multirow{2}[0]{*}{\ding{51}} \\
    \hline\specialrule{0em}{1.5pt}{1pt}
    \multirow{2}[0]{*}{9} & Go to Amazon and add the cheapest charger into the shopping cart. & \multirow{2}[0]{*}{4/7 (9)} & \multirow{2}[0]{*}{\ding{55}} & \multirow{2}[0]{*}{5/5 (9)} & \multirow{2}[0]{*}{\ding{55}} & \multirow{2}[0]{*}{9/9 (9)} & \multirow{2}[0]{*}{\ding{51}} \\
    \hline\specialrule{0em}{1.5pt}{1pt}
    \multirow{2}[0]{*}{10} & Add my Costco's loyalty card number \{card\_num\} in the website \{web\_url\}. & \multirow{2}[0]{*}{2/6 (6)} & \multirow{2}[0]{*}{\ding{55}} & \multirow{2}[0]{*}{3/3 (6)} & \multirow{2}[0]{*}{\ding{55}} & \multirow{2}[0]{*}{6/7 (6)} & \multirow{2}[0]{*}{\ding{51}} \\
    \hline\specialrule{0em}{1.5pt}{1pt}
    \multirow{2}[0]{*}{11} & Create a meeting at 2023/04/15 14:00-14:30 in Outlook. & \multirow{2}[0]{*}{3/6 (8)} & \multirow{2}[0]{*}{\ding{55}} & \multirow{2}[0]{*}{3/3 (8)} & \multirow{2}[0]{*}{\ding{55}} & \multirow{2}[0]{*}{3/3 (8)} & \multirow{2}[0]{*}{\ding{55}} \\
    \hline\specialrule{0em}{1.5pt}{1pt}
    \multirow{2}[0]{*}{12} & Search the Cpython repo and download its zip file in github.com. & \multirow{2}[0]{*}{5/7 (8)} & \multirow{2}[0]{*}{\ding{55}} & \multirow{2}[0]{*}{6/6 (8)} & \multirow{2}[0]{*}{\ding{55}} & \multirow{2}[0]{*}{6/6 (8)} & \multirow{2}[0]{*}{\ding{55}} \\
    \bottomrule
    \end{tabular}%
  }
  \label{tab:case_study}%
\end{table}%

\begin{figure}[ht]
	\begin{center}
		\includegraphics[width=0.95\textwidth]{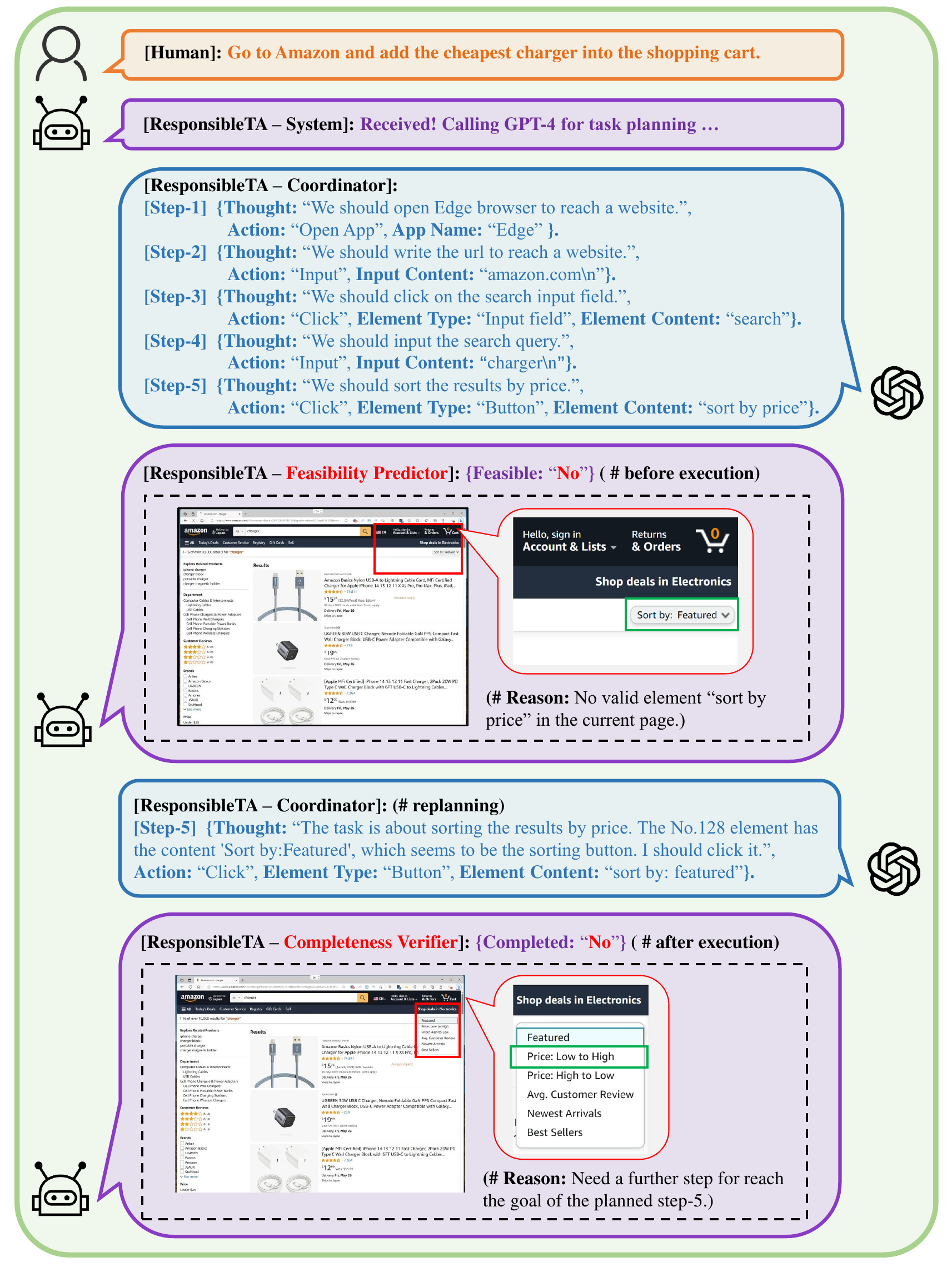} 
	\end{center}
% 	\caption{Detailed case study on how our proposed feasibility predictor and completeness verifier in \ours~remedy the failure case to achieve success. We take the No.9 task in Table \ref{tab:case_study} as an example, and omit its $6$-th to $9$-th steps for brevity.}
    \caption{Detailed case study about how our proposed feasibility predictor and completeness verifier in \ours~remedy the failure case to achieve success on the No.9 task in Table \ref{tab:case_study}. The $6$-th to $9$-th steps are omitted for brevity. GPT-4 \cite{GPT4} is used as the LLM-based coordinator.}
	\label{fig:case_study}
\end{figure}

\subsection{Case Study and Demonstration}

We analyze the performance and behaviors of our proposed \ours~via real-world case study on 12 tasks, considering that there are no fledged benchmarks in this research field yet. Besides, we conduct ablation study on these tasks by comparing \textit{Baseline}, \textit{Baseline+Fea.} and \textit{Baseline+Fea.+Com.} where \textit{Baseline+Fea.+Com.} is the complete version of \ours. Their configurations are detailed in the caption of Table \ref{tab:case_study}. The security predictor is 
installed in all models.
% and omitted here for conciseness.

\paragraph{Ablation study and analysis.} We report the specific completion progress and the final success status over all 12 real-world tasks in Table \ref{tab:case_study}. As for the success ratio, we can find that \textit{Baseline} does not reach the final success on 9 out of 12 tasks. For these 9 tasks, 5 of them are successfully remedied by the feasibility predictor on its own. On top of it, the completeness verifier help turn 2 of them to the final success in addition. It can be seen that the feasibility predictor and the completeness verifier in \ours~can significantly improve the success ratio of task automation by providing feedbacks for LLM-based coordinator so that it can replan timely. \textit{How our \ours~achieves this} will be elaborated in detail in the subsequent text. 

Besides the benefits of improving the success rate, our proposed \ours~can effectively reduce the number of invalid instruction executions thanks to its feasibility predictor. This conclusion is drawn by comparing the valid steps and total execution steps for \textit{Baseline} and \textit{Baseline+Fea.} models. Thus, our \ours~can avoid the risks when executing those invalid steps.

Moreover, we further verify the role of \ours~on security protection by comparing using placeholders and real user information on the No.7 and No.10 tasks. We experimentally find that launching the security protector neither affects the task success rate nor increases the number of invalid executions on these two cases for all three models. It can effectively obviate the need of uploading user-sensitive information to the cloud.

\paragraph{How \ours~remedies failure cases?} Here, we take a close look into a specific case for detailed analysis on how \ours~remedies a failure case to achieve success in the end. We illustrate specific execution processes of the first 4 steps for the No.9 task in Figure \ref{fig:case_study}. The execution for remaining steps and the corresponding prompts are omitted here for brevity. As shown in Figure \ref{fig:case_study}, the LLM-based coordinator generates feasible low-level commands for the first 4 steps, and the corresponding execution is smooth for these steps. For the $5$-th step, the coordinator originally thinks we should click the button with the content of ``sort by price''. It is seemingly a reasonable command but does not match the current web page in fact, because there is no matched button. At this point, the \textit{Baseline} model executes this infeasible command in a seemingly random manner, leading to 4 valid steps in total. When \ours~has the feasibility predictor, this module will return ``infeasible'' signal to the coordinator so that it can replan upon additionally provided screen page information. After the replanning, the coordinator gives a feasible command, \ieno, click the ``Sort by: Featured'' button, and yields one more valid step. Despite this, \textit{Baseline+Fea.} still fails for the next step since it does not realize that another click operation is further needed to achieve the goal of sorting by price. This can be effectively remedied by the complete version of \ours~thanks to its completeness verifier. With the assist of this module, \ours~add such click operation for the ``Price: Low to High'' button before executing subsequent steps via another replanning. In this way, it adjusts the planned actions timely and contributes to the eventual success.

\section{Limitation}

We have to admit that it is difficult, even impossible, to cover all aspects of a newly devised system in a single academic paper. This paper makes contributions and delivers insights from the perspectives of framework design and technical paradigm comparison. Its remaining limitations lie in two aspects: 1) benchmark construction; 2) specific model design/iteration aimed at pursuing high performance for this new domain. We here call on the community to address them with our work as a start point.

\section{Conclusion and Broader Impact}

In this paper, we present a fundamental multi-modal framework, \ours, for empowering LLMs as responsible task automators. In specific, we enhance LLMs with three core capabilities, \ieno, feasibility prediction, completeness verification and security protection. Moreover, we propose and compare different technical paradigms for implementing these core functionalities.
% , and evaluate their effectiveness via quantitative extensive case study. 
We experimentally observe that domain-specific models deliver superior performance compared to prompt engineering based solutions on their implementations, while the later one does not require collecting domain-specific data for model training. Besides, we also demonstrate the effectiveness of our proposed \ours~and provide explanations on how our \ours~works intuitively through case study. As for the broader impacts, we hope our work can inspire more excellent works on the related benchmark construction, method design and functionality extension on top of this work in the future.

% Considering this is a very emerging research field, we make the first attempt to devise a multi-modal fundamental framework for responsible task automation, and hope it can inspire more excellent works on benchmark construction, method design and functionality extension on top of our this work in the future. 
% The limitations are discussed in detail in the supplementary.

%%%%%%%%%%%%%%%%%%%%%%%%%%%%%%%%%%%%%%%%%%%%%%%%%%%%%%%%%%%%

\bibliography{neurips_2023}
\bibliographystyle{plain}

\clearpage
\appendix

\textbf{[Supplementary Material]}
\section{Further Model Introduction}

We elaborate the three core modules in our proposed \ours, \ieno, the feasibility predictor, completeness verifier and security protector, with their modeling in Sec.3 and their used datasets in Sec.4 in the main body of our paper. Besides these three core modules, in \ours, we also train a domain-specific command executor and a screen parsing model. The domain-specific command executor aims to locate the target UI element by predicting its spatial coordinates for automating clicking or typing operations in accordance with given commands. And the screen parsing model converts a given screenshot into a series of element-wise descriptions in linguistic form, which plays the role of inputting the information of a screenshot in linguistic form to the LLM-based coordinator in two scenarios: 1) when replanning; 2) when employing the prompt engineering based paradigms for implementing the feasibility predictor or completeness verifier, as proposed. This model is needed in consideration to that most of LLMs have not developed or released their visual input APIs currently. These two modules are not the highlights of this work. We thus detail them in this supplementary material.

\subsection{Domain-specific Executor}
\label{sec:ds_specific}

The domain-specific executor is a multimodal model that accepts both a screenshot and a command as its inputs. It is analogous to the domain-specific model-based paradigm introduced for implementing the feasibility predictor or completeness verifier in the main text. Inspired by Pix2Seq modeling \cite{chen2021pix2seq,chen2022unified}, we employ the same architecture design for this model with that of the feasibility predictor as illustrated in Figure 2 of the main text. It requires different instantiations for the structured output format, \ieno, ``\textit{<task\_prompt> \{results\} </task\_prompt>}''. In this model, the ``\textit{<task\_prompt>}'' and ``\textit{</task\_prompt>}'' are instantiated by ``\textit{<locate\_element>}'' and ``\textit{</locate\_element>}'', respectively. And the ``\textit{\{results\}}'' is organized as ``\textit{<x\_min> \{$x_{min}$\} </x\_min> <y\_min> \{$y_{min}$\} </y\_min> <x\_max> \{$x_{max}$\} </x\_max> <y\_max> \{$y_{max}$\} </y\_max>}'' wherein [$x_{min}$, $y_{min}$, $x_{max}$, $y_{max}$] denotes the coordinates of the top-left and bottom-right points of the bounding box corresponding to the target UI element. It achieves 0.51 mIoU for locating the target UI elements in given commands.

\subsection{Screen Parsing Model}
\label{sec:scree_parser}

% As mentioned in our main body, except for the coordinator, the LLM is also used in replanning, prompt engineering based feasibility predictor and completeness verifier, where we provide the UI screenshot to the LLM and make it to replan the task, predict feasibility or completeness. Since the most of advanced LLMs (e.g., ChatGPT, GPT-4) have not integrated or released their input interfaces for other modalities except for language, we have to train a screen parsing model to ``translate'' the UI screenshot into UI element based linguistical form to communicate with the LLM. 

% We formulate the UI element as the basic unit that makes up the UI, which consists of four attributes: \textit{index}, \textit{text content}, \textit{location} and \textit{type}. The \textit{index} of a element is the unique identifier of this element in the UI. The \textit{text content} is the specific content of a element. The \textit{location} of a element is represented by the two-dimensional coordinates of its center point. For the \textit{type} of a element, we categorize it into one of \textit{button}, \textit{input} and \textit{icon}. 

The screen parsing model aims to detect all UI elements in a given screenshot and recognize their attributes, \ieno, the location, text content, and type. Regarding the type attribute, we categorize each UI element into one of \textit{button}, \textit{input}, and \textit{icon}. This model is a mixture of expert models including element detector, text detector, text recognizer, and icon recognizer. For a given UI screenshot, the element detector first locates all UI elements. Then, for button and input elements, the text detector locates their text regions when texts are available, and the text recognizer extracts their text contents. For icon elements, icon recognizer recognizes their categories as the text contents.
% The goal of screen parsing model is to localize every UI element and recognize its four attributes in the screenshot. Our screen parsing model consists of four models including element detector, text detector, text recognizer and icon recognizer. Given a UI screenshot as input, the element detector first localizes the \textit{button}, \textit{input} and \textit{icon} elements. Then the text detector and recognizer localize and recognize text regions and their corresponding content respectively. For button and input elements, according to the IoU(intersection over union) with them, the texts are subsequently assigned as the contents of corresponding elements. For icon elements, icon recognizer recognizes the categories of the icons as the element contents.
Specifically, for element detector, we adopt RTMDet \cite{lyu2022rtmdet}-style architecture with ShuffleNetv2-1.0x \cite{ma2018shufflenet} backbone. 
% To train the element detector, we build a dataset based on publicly available web pages and windows apps. We capture screenshots of these UIs and extract types and bounding boxes of leaf nodes from their tree-like metadata, \ieno, DOM \cite{DOM} and UIA \cite{UIA}, as element annotations. Finally an element detection dataset including around 1.5 million images and corresponding element annotations is built, where around 1.2 million samples are used for training and another 0.3 million are used for test. 
It achieves 0.710 mAP on the test set introduced as follows. 
% Our trained element detector achieves 0.710 mAP on our test dataset.
For text detector and text recognizer, we employ the off-the-shelf models from PaddleOCRv3 \cite{li2022ppocrv3}. 
% For text detector and text recognizer, we adopt corresponding off-the-shelf models from PaddleOCRv3 \cite{li2022ppocrv3}.
For icon recognition, we use ShuffleNetv2-1.0x \cite{ma2018shufflenet} as the backbone of the icon classifier and use a fully connected layer as the classification head. 
% Similar to element detector, to train our icon recognizer, we also build an icon classification dataset based on Rico dataset\cite{Liu:2018:LDS:3242587.3242650} containing 14,043 icon images and corresponding annotations for 14 frequently used icon categories, where training split contains 12,637 samples and test split contains 1,405 samples. 
Our icon recognizer achieves 95.7\% averaged accuracy on the test set.

\section{Further Dataset Introduction}

We elaborate the datasets used for domain-specific feasibility predictor and completeness verifier in the main text. In this section, we further introduce the data for aforementioned domain-specific executor and screen parsing model.

The dataset for domain-specific executor consists of all feasible screenshot-instruction pairs from the feasibility prediction dataset introduced in Sec.4.1 of the main text. Its training split contains 0.5M samples from 38K desktop screenshots, and its testing split contains 27K samples from 2K desktop screenshots. For the element detector in the screen parsing model, we collect a dataset upon publicly available web pages and windows apps, comprising around 1.5M screenshots with 1.2M of them as the training split and 0.3M of them as the testing split. For these data, we obtain the annotations of UI elements, \ieno, their types and bounding boxes, from their tree-structure metadata, \ieno, DOM \cite{DOM} and UIA \cite{UIA}. Only leaf nodes are used. For the icon classifier in the screen parsing model, we build a dataset based on a public one (Rico \cite{Liu:2018:LDS:3242587.3242650}), which contains 14,043 icon images with 14 frequently used icon categories. Its training split contains 12,637 samples while its test split contains 1,405 samples. 

\begin{figure}[!t]%[h]
	\begin{center}
		\includegraphics[width=0.95\textwidth]{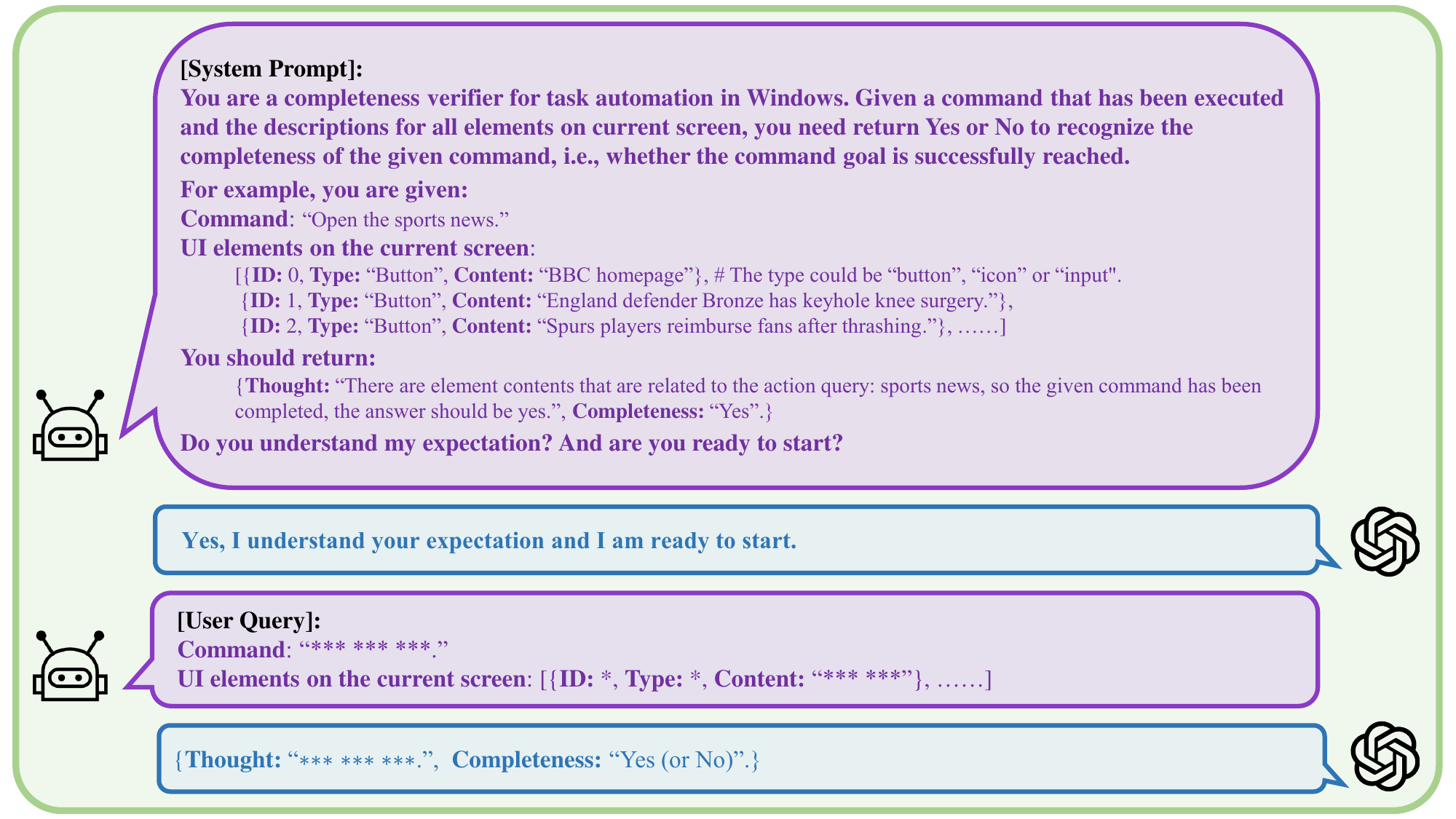} 
	\end{center}
	\caption{Illustration of our prompt engineering based paradigm for implementing the completeness verifier in our proposed \ours.}
	\label{fig:model_llmcv}
\end{figure}

\section{More Implementation Details}

\subsection{Training Details}

As introduced, in \ours, the feasibility predictor, completeness verifier and domain-specific executor share the same model architecture design as shown in Figure 3 of the main text. For this architecture, we employ Swin Transformer \cite{liu2021swin} and BART model \cite{lewis2019bart} as its vision encoder and language decoder, respectively. For all of them, we first pretrain the entire model on document understanding tasks introduced in \cite{kim2022ocr} and then finetune it on those datasets for feasibility prediction, completeness verification and command execution. Unless specifically stated, we perform the finetuning on each task for 20 epochs using 8 NVIDIA V100 GPUs, with a batch size of 2 on each GPU card. The height and width of screenshots are resized to 960 and 1280, respectively. We use the Adam optimizer \cite{kingma2014adam} and set the initial learning rate to be $1\!\times\!10^{-4}$. Besides, we apply a cosine learning rate annealing schedule and a gradient clipping technique with the maximum gradient norm of 1.0.

\subsection{Prompt Design Details}

Similar to the proposed paradigms for implementing the feasibility predictor, we also introduce two analogical paradigms for implementing the completeness verifier in our proposed \ours. Regarding the prompt engineering based paradigm, we detail its related prompt design as illustrated in Figure \ref{fig:model_llmcv} for clearer introduction and better reproducibility.

\section{More Experiment Results}

\begin{figure*}[!htbp]%[h]
	\begin{center}
		\includegraphics[width=\textwidth]{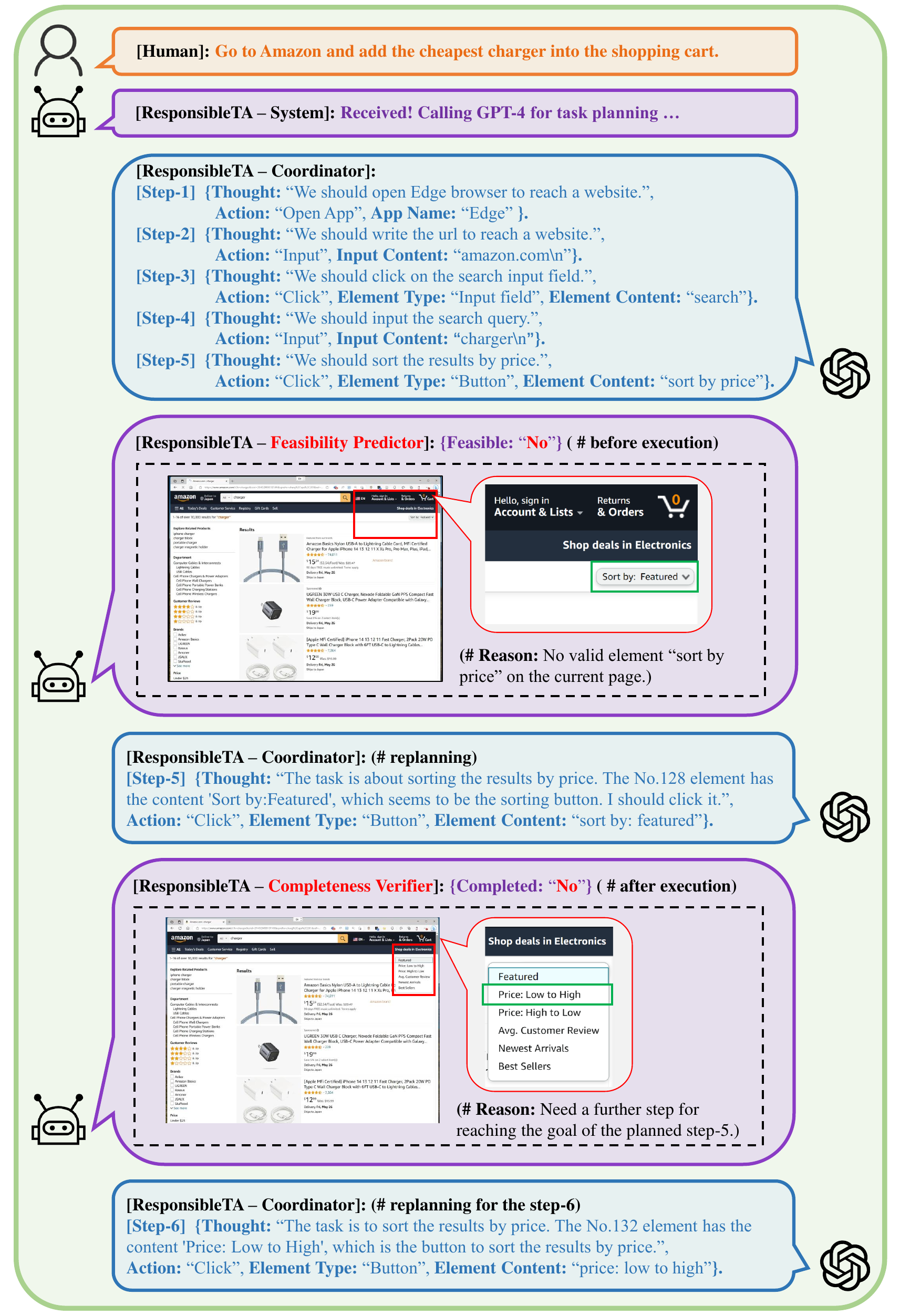} 
	\end{center}
	\caption{Illustration of a successful case (\ieno, the No.9 task in Table 2 of our main text) about how our proposed feasibility predictor and completeness verifier in \ours~remedy the failure case to achieve success. This figure depicts the part-1 of its complete version, \ieno, from the $1$-th step to the $6$-th step. GPT-4 \cite{GPT4} is used as the LLM-based coordinator.}
	\label{fig:complete_part1}
\end{figure*}

In Figure 4 of the main text, we have depicted the automation process of the first five steps on a specific task (\ieno, Task 9 in Table 2 of our main text) to show how our proposed feasibility predictor and completeness verifier pla their roles in turning an originally failed case into a successful one. Here, in Sec.\ref{sec:successful_case}, we provide its complete version with its part-1 (from the beginning to the $6$-th step) illustrated in Figure \ref{fig:complete_part1} and its part-2 (from the $7$-th step to the end) illustrated in Figure \ref{fig:complete_part2}. Furthermore, we provide a failure case (illustrated in Figure \ref{fig:case_failure}) and its analysis in Sec.\ref{sec:failure_case}. 

\subsection{A Successful Case and Its Analysis}
\label{sec:successful_case}

\begin{figure}[!htbp]%[h]
	\begin{center}
		\includegraphics[width=\textwidth]{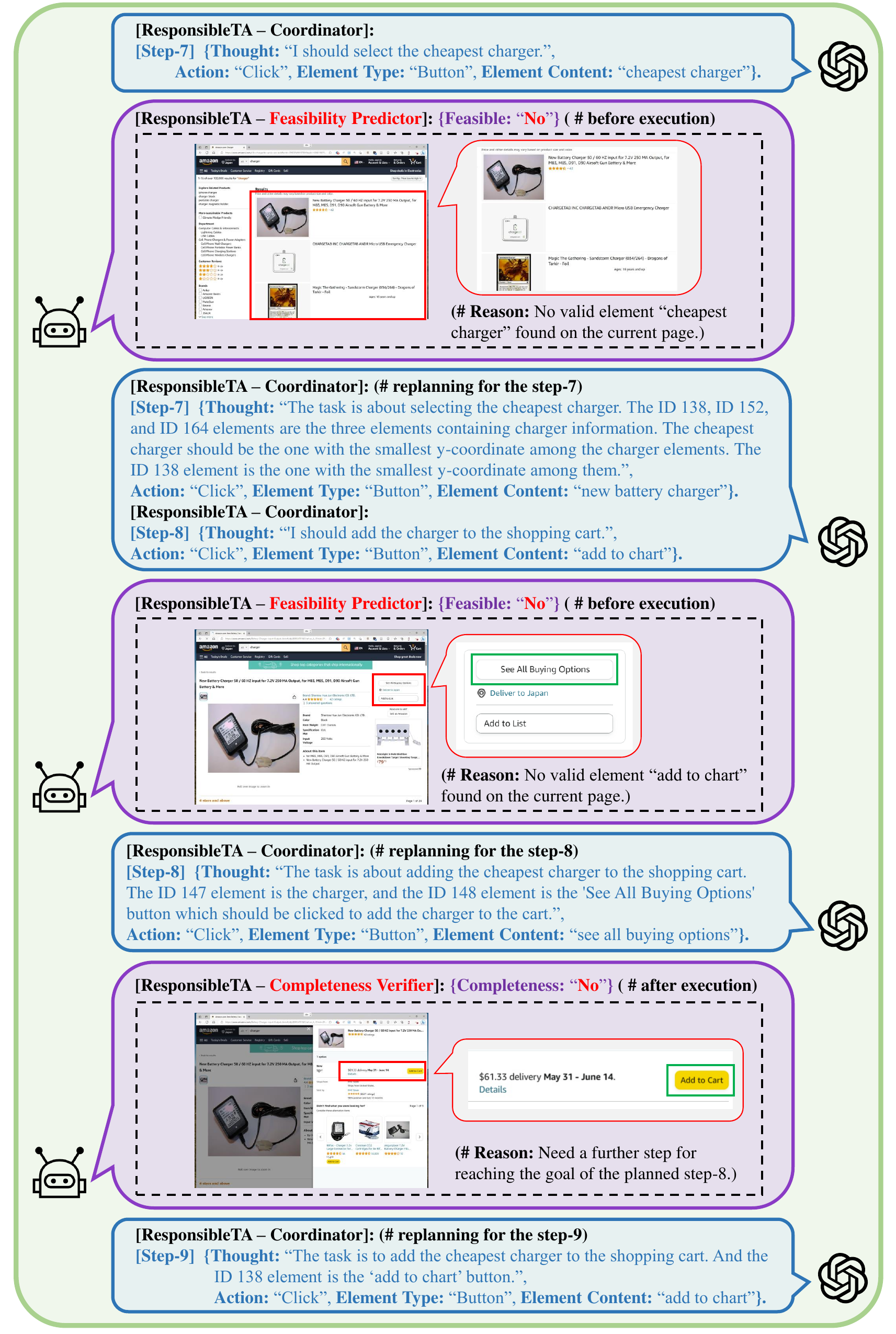} 
	\end{center}
	\caption{Continuing from the Figure \ref{fig:complete_part1} in this supplementary material, this figure depicts the part-2 of the complete automation process of No.9 task, \ieno, from the $7$-th step to the end.}
	\label{fig:complete_part2}
\end{figure}

Note that the in-depth analysis for the part-1 of this case is in Sec.4.3 of the main text. We provide the detailed analysis regarding its part-2 here. As shown in Figure \ref{fig:complete_part2}, the GPT-4 based coordinator in \ours~originally plans to click the button with the content of ``cheapest charger''. However, in the real web page, there is no matched element on the current page. At this time, the feasibility predictor considers this planned command as an infeasible one before execution, and asks the coordinator for a replanned command upon the information of the current page. Then, the coordinator thinks we should click the element containing charge information with the smallest $y$-coordinate so that this step is correctly processed. The coordinator plans for the next step, \ieno, adding the selected item to the shopping cart. It gives an infeasible command again since there is no ``add to chart'' item on the current page. This planned goal requires two execution operations to be completed in actual. With the help of the feasibility predictor and completeness verifier, our \ours~utimately achieves the purpose of adding the item to the shopping cart by clicking the ``See All Buying Options'' button followed by the ``Add to Cart'' button. As such, the human instruction ``\textit{Go to Amazon and add the cheapest charger into the shopping cart.}'' is successfully automated.

From the detailed analysis of this case, we can intuitively understand the functions of the feasibility predictor and completeness verifier in \ours. In specific, the feasibility predictor can intercept unreasonably planned commands. And the completeness verifier checks whether the actual executed operations have achieved the intended goals step-by-step. They serve as a double guarantee for \ours~to responsibly achieve task automation before and after command execution, by providing feedbacks for the coordinator so that it can perform replanning timely.

\subsection{A Failure Case and Its Analysis}
\label{sec:failure_case}

\begin{figure*}[!htbp]%[h]
	\begin{center}
		\includegraphics[width=\textwidth]{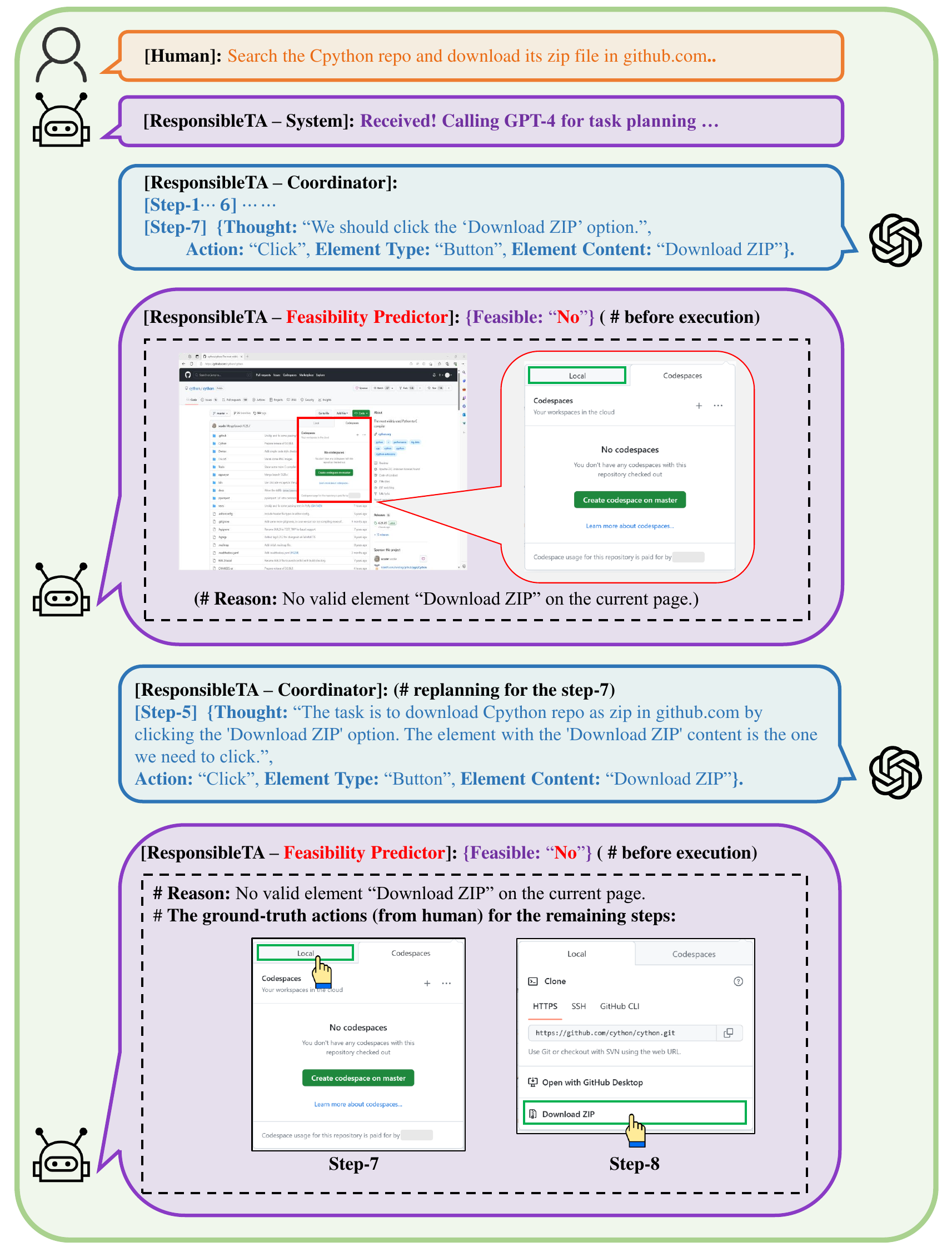} 
	\end{center}
	\caption{Illustration of a failure case (\ieno, the No.12 task in Table 2 of our main text). The first six steps are omitted in this figure for the brevity. GPT-4 \cite{GPT4} is used as the LLM-based coordinator.}
	\label{fig:case_failure}
\end{figure*}

We describe a failure case (\ieno, the No.12 task in Table 2 of our main text) that the feasibility predictor and completeness verifier cannot remedy, as illustrated in Figure \ref{fig:case_failure}. This failure happens in automating the human instruction ``\textit{Search the Cpython repo and download its zip file in github.com.}''. In most GitHub repositories, we can achieve the download purpose by directly clicking the ``Download ZIP'' button. However, in some GitHub repositories, such as the one in our illustrated failure case, the ``Download ZIP'' button is hidden in a secondary menu. In this case, we are required to complete the download of the ZIP file through two operations: first clicking the ``Local'' button, and then clicking the ``Download ZIP'' button. As shown in Figure \ref{fig:case_failure}, although our proposed module accurately detects that the commands given by the coordinator are infeasible, the coordinator has not been able to provide correct and feasible commands via its replanning. This task is ultimately terminated when the preset maximum number of replanning attempts is reached. This failure case implies that the knowledge of current LLMs is generic but may not be perfect, and there is room for our proposed \ours~to become more powerful as the capabilities of LLMs improve.

% \section{Demo Video}

% We provide a demo video for the entire automation process of the case introduced in Sec.\ref{sec:successful_case} in this link: \url{https://anonymous.4open.science/r/responsible_task_automation_demo_video-C9D9/}.

% \section{Typo Correction}

% We incorporated an outdated version of Figure 4 in our main text by a mistake. Please allow us to make minor revisions for this figure here to avoid potential confusions. The human instruction should be ``Go to Amazon and add \textit{the cheapest chart} into the shopping cart.'', corresponding to the No.9 task in Table 2 of our main text. The system name ``ReliableTA'' should be ``\ours''. The ``reach'' in the annotation of completeness verifier's response should be ``reaching''. Please kindly note that the Figure \ref{fig:complete_part1} in this supplementary material could be viewed as a updated version of the Figure 4 in our main text. The ``committee'' in Line314 should be ``community''. Sorry for these typos.

\end{document}